\documentclass[sigconf,nonacm]{acmart}

\usepackage{amsmath}
\usepackage{booktabs}

\newcommand{\pbox}[1]{\par\smallskip\noindent\fbox{\begin{minipage}{0.93\columnwidth}\small #1\end{minipage}}\par\smallskip}

\begin{document}

\title{Is Your Neighborhood Safe? Place-based Stigma in Large Language Models' Urban Safety Judgments}

\author{Huy Nguyen}
\email{huynguyen23@augustana.edu}
\affiliation{%
  \institution{Augustana College}
  \city{Rock Island}
  \state{Illinois}
  \country{USA}
}

\author{Yue Lin}
\email{linyue@illinois.edu}
\affiliation{%
  \institution{University of Illinois Urbana-Champaign}
  \city{Urbana}
  \state{Illinois}
  \country{USA}
}

\renewcommand{\shortauthors}{Nguyen and Lin}

\begin{abstract}
Large language models are increasingly used to inform safety decisions in cities, such as where it is safe to walk, rent, or travel. We ask whether such judgments track \emph{measured risk} or the patterns attached to an urban neighborhood's \emph{name}. We probe seven instruct-tuned models to rate night-time walking safety under three conditions that dissociate the name from geography, consisting of coordinates-only, name-only, and name$+$coordinates, over 186 neighborhoods in Los Angeles and Chicago, joined to recorded violent crime and American Community Survey demographics.
First, ratings are nearly flat under coordinates for six of the seven models while the name carries essentially all between-neighborhood variation, and that variation is moderately calibrated to violent crime in every model; only at frontier scale does the coordinate channel itself carry appreciable amplitude.
Second, the name lowers the safety rating more for neighborhoods with a higher share of the locally dominant marginalized group (percent Black in Chicago, percent Hispanic in Los Angeles);
the name effect tracks demographic share in all seven models in both cities. Whether it exceeds what crime justifies can only be tested where the two are separable: in Chicago percent Black and crime correlate at $r=0.75$, too closely to tell apart, while in Los Angeles the demographic effect survives controlling for crime and income in all seven models and is confirmed at the item level by crime-matched pairs. Addressing the worry that recorded crime reflects policing, an enforcement-elasticity decomposition shows the over-caution tracks near-fully-reported homicide rather than discretionary, deployment-driven offences, and this replicates in Los Angeles where the two are statistically independent. Third, the effect scales with geographic knowledge rather than being gated by it: models that better distinguish a city's real neighborhoods apply more demographic stereotype to them.
Crucially, the name carries genuine crime signal and demographic stereotype in a single channel: stripping the name removes the bias and the accuracy together, so the failure cannot be fixed by hiding the name. We discuss implications for deploying LLMs in advice and decision-support settings.
\end{abstract}



\maketitle

\section{Introduction}
Large language models (LLMs) have increasingly become a key driver of the vision of intelligent cities through applications in search, chat assistants, travel tools, and recommendation systems, and now power agentic artificial intelligence (AI) systems for autonomous decision-making that could shape the socio-physical environments of cities \citep{xu2025agentic,li2025language,han2025large}. Among these applications, questions of safety are especially urgent, given persistent public safety and crime concerns in certain urban environments. Recent studies have begun to investigate how LLMs assess and simulate urban safety, either on their own or with additional data such as street-view imagery \citep{han2026enhancing,bougie2025citysim}.

Given the growing use of LLMs in urban safety applications, this paper asks whether LLM judgments match measured risk, or reflect existing racial-spatial patterns of neighborhood stigma. Studies have shown that neighborhoods with higher shares of Black residents are more likely to trigger assumptions of high crime and disorder, even when measured crime rates are not significantly higher \citep{quillian2001,sampson2004seeing}. This stigma also attaches to names in informal discourse. The names of certain neighborhoods become shorthand for places ``you just don't go,'' with danger presumed independent of measured risk \citep{jones2018you}. This pattern extends to how neighborhood names are represented in digital spaces. A recent study found that online housing listings included the names of neighborhoods with more Black residents less often than those of white neighborhoods \citep{schachter2024s}. Our goal is to examine whether, and how, this pattern carries over into LLM judgments of urban safety that may further penalize neighborhoods of marginalized populations more than what recorded crime justifies.

We disentangle these mechanisms with a controlled experimental design across neighborhoods in Chicago and Los Angeles. For each neighborhood, we collect night-time walking-safety ratings under conditions that vary only what identifying information the model receives: coordinates-only, name-only, and name$+$coordinates. Because coordinates specify a location without revealing its name, comparing the coordinates-only condition to the name-bearing conditions isolates the name's effect on the rating. We run this comparison across seven models (a primary set of four, plus three larger models for the scaling analyses) and two cities, Chicago and Los Angeles, and align every rating with recorded violent crime and census demographics.

\paragraph{Contributions.}
\begin{itemize}
    \item A condition-ablation probe that separates a neighborhood \emph{name} from its \emph{coordinates} and measured \emph{risk}, applied to seven LLMs (a primary set of four, plus Qwen2.5-14B, GPT-4o, and DeepSeek V4 for the scaling analyses) over 186 neighborhoods in two cities. Coordinates alone barely affect the rating, and neighborhood names account for nearly all the variation.
    \item Evidence that the name depresses safety ratings more for higher-percent-Black (Chicago) and higher-percent-Hispanic (Los Angeles) neighborhoods, in all seven models in both cities. Net of measured crime and income the effect is identified in Los Angeles, where all seven models show it and three independent estimators agree: the level regression, the name ablation, and crime-matched pairs. Chicago cannot separate the two, since percent Black and crime correlate at $r=0.75$ there. The pattern is robust to prompt wording, rating scale, and two measures of name prominence (Wikipedia pageviews and pretraining-corpus frequency).
    \item An enforcement-elasticity test that addresses the standard objection that recorded crime reflects policing: sorting incidents from near-fully-reported homicide to deployment-driven discretionary offenses, the over-caution loads on the reported end and gains essentially nothing from the discretionary end net of homicide, and this replicates in Los Angeles, where the two ends are statistically independent.
    \item Evidence that the demographic effect \emph{scales with} geographic knowledge rather than being gated by it: two independent knowledge measures both predict the size of the name effect across seven models (region accuracy $\rho=-0.94$, $p=0.002$ in Los Angeles; recognition $d'$ $\rho=-0.71$ and $-0.75$).

    \item Evidence that the coordinate channel is not uniformly inert: negligible for the open models (SD $0.01$--$0.15$), it reaches $0.67$--$0.91$ at frontier scale, so the result that the name carries everything is itself scale-dependent.

    \item A decomposition showing that the name channel carries real crime signal and demographic stereotype simultaneously, so the bias cannot be removed by hiding the name without also removing the accuracy.

\end{itemize}

\section{Related Work}
\paragraph{Perceived neighborhood safety and racial stigma.}
Studies have documented racial-spatial patterns in perceived neighborhood safety. Perceived neighborhood crime rises with the share of young Black men, even after accounting for independent measures of actual crime, across Chicago, Seattle, and Baltimore \citep{quillian2001}. In another study, perceived disorder rises with minority and poverty concentration after accounting for independently observed disorder, across roughly 500 Chicago block groups, framed as self-reinforcing neighborhood stigma \citep{sampson2004}. The mechanism is one of stereotype amplification \citep{quillian2010}. This pattern of neighborhood stigmatization extends to data representations in the digital space \citep{schachter2024s}. We ask whether a language model, trained on data from the digital space, reproduces the place misrepresentation.

\paragraph{Bias in learned language representations.}
A long line of work documents social bias in learned language representations within AI systems. Early studies found geometric bias in static word embeddings \citep{bolukbasi2016} and human-like association biases recovered from corpora \citep{caliskan2017}. More recent work extends this to bias and stereotype in modern language generation and large models \citep{sheng2019,nangia2020,abid2021}, with broad analyses of scale warning that fluent models can encode and amplify social stereotypes \citep{bender2021,weidinger2021}. Our study extends this line of work with an experimental design built around neighborhood names in a safety rating task, to measure how bias manifests in geospatial and urban decision-making.

\paragraph{Geographic bias in LLMs.}
LLMs can produce geographically uneven and biased information about regions \citep{manvi2024}, and such bias may be amplified with model size \citep{godey2024,bhandari2023}. This bias also has a sociodemographic dimension. One study found that LLM responses to local-knowledge prompts show an ``urban advantage,'' with rural areas systematically underrepresented and described with lower semantic depth \citep{gao2026your}. A separate study, in the context of environmental justice, found that LLMs performed better for inland and more urbanized counties than for rural and low-income areas \citep{kim2024exploring}. Related neighborhood-level research also documents systematic misrepresentation of neighborhoods with larger shares of racial and ethnic minority residents \citep{lin2025would}.

\section{Data}
\subsection{Neighborhood Units and Crime Data}
We study two cities. For \textbf{Chicago}, we use the 77 official community areas and recorded incidents from the City's open crime data, consisting of crime records from January 2020 through April 2025. For \textbf{Los Angeles}, we use the 109 residential City-of-LA neighborhoods from the ``Mapping L.A.'' boundaries. Mapping L.A. also designates parks, basins, and similar non-residential
polygons as neighborhoods. These record crime incidents but have near-zero resident populations, so their per-capita crime rates are unstable. We therefore retain only polygons with at least 500 residents, in which the smallest we kept has $2{,}317$. This also keeps our design focused on places where ``walking alone at night'' has its ordinary residential meaning. Los Angeles Police Department incident data contains crime records from January 2020 through December 2024, spatially joining each incident to its neighborhood. In both cities, we compute, for each neighborhood, a violent-crime count. In Chicago this is the exact set of \texttt{Primary Type} values \{homicide, criminal sexual assault, robbery, battery, assault, kidnapping, intimidation, human trafficking\} ($393{,}927$ incidents, $33.9\%$ of the total). In Los Angeles it is the 25 \texttt{Crm Cd Desc} values covering the same concepts: homicide, manslaughter, robbery, assault, battery, rape and other sexual offenses, kidnapping, human trafficking, and intimate-partner offenses ($247{,}151$ incidents, $24.9\%$). The two cities use different reporting schemas, so the definitions cannot be made identical. Because every analysis compares neighborhoods only \emph{within} a city, conceptual alignment is what the design requires, not exact equivalence. The second quantity is the mean latitude/longitude of its recorded incidents, used as the anchor point in the coordinates-only condition. We also use homicide rate as an alternative measure that is less sensitive to enforcement bias, since homicide is less discretionarily recorded than lower-level offenses (Section~\ref{sec:tiers}).

\subsection{Demographics}
We aggregate American Community Survey (ACS) 5-year tract estimates, including total population, households, aggregate household income, and race and ethnicity shares (non-Hispanic White, non-Hispanic Black, and Hispanic), into our neighborhood data for both cities. Mean household income is aggregate income divided by households. We express risk as violent crimes per 1,000 residents.

The two cities differ in how racial and ethnic groups are distributed across neighborhoods. In Chicago, percent Black ranges up to roughly 96\% in certain neighborhoods and has been found to be strongly correlated with the violent-crime rate. In Los Angeles, percent Black is narrower in range (maximum $\approx$66\%), while percent Hispanic reaches up to 92\% in some neighborhoods. We therefore analyze each city on its salient demographic axis, and report both axes where informative.

\begin{figure*}[tp]
\centering
\includegraphics[width=0.8\textwidth]{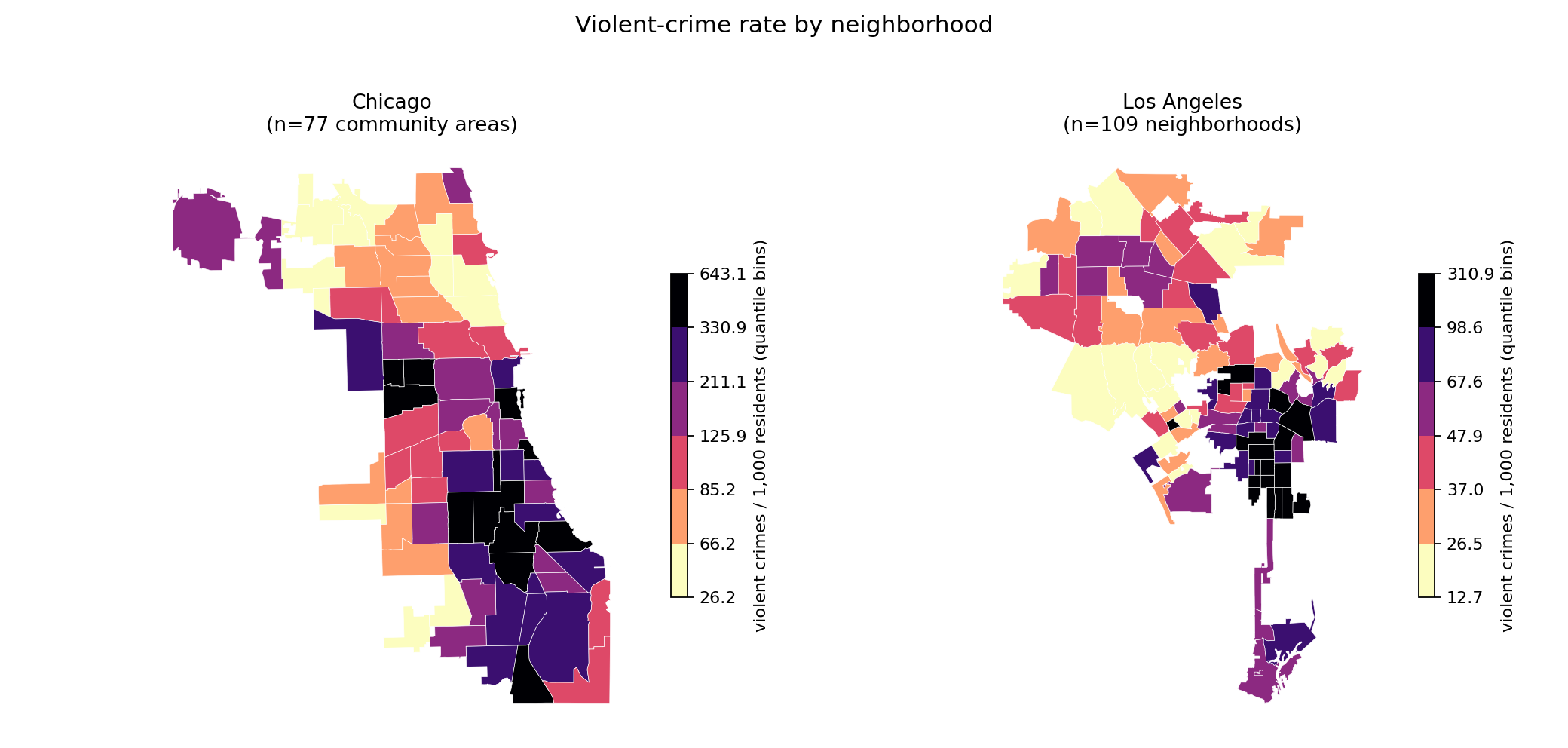}\\[2pt]
\includegraphics[width=0.8\textwidth]{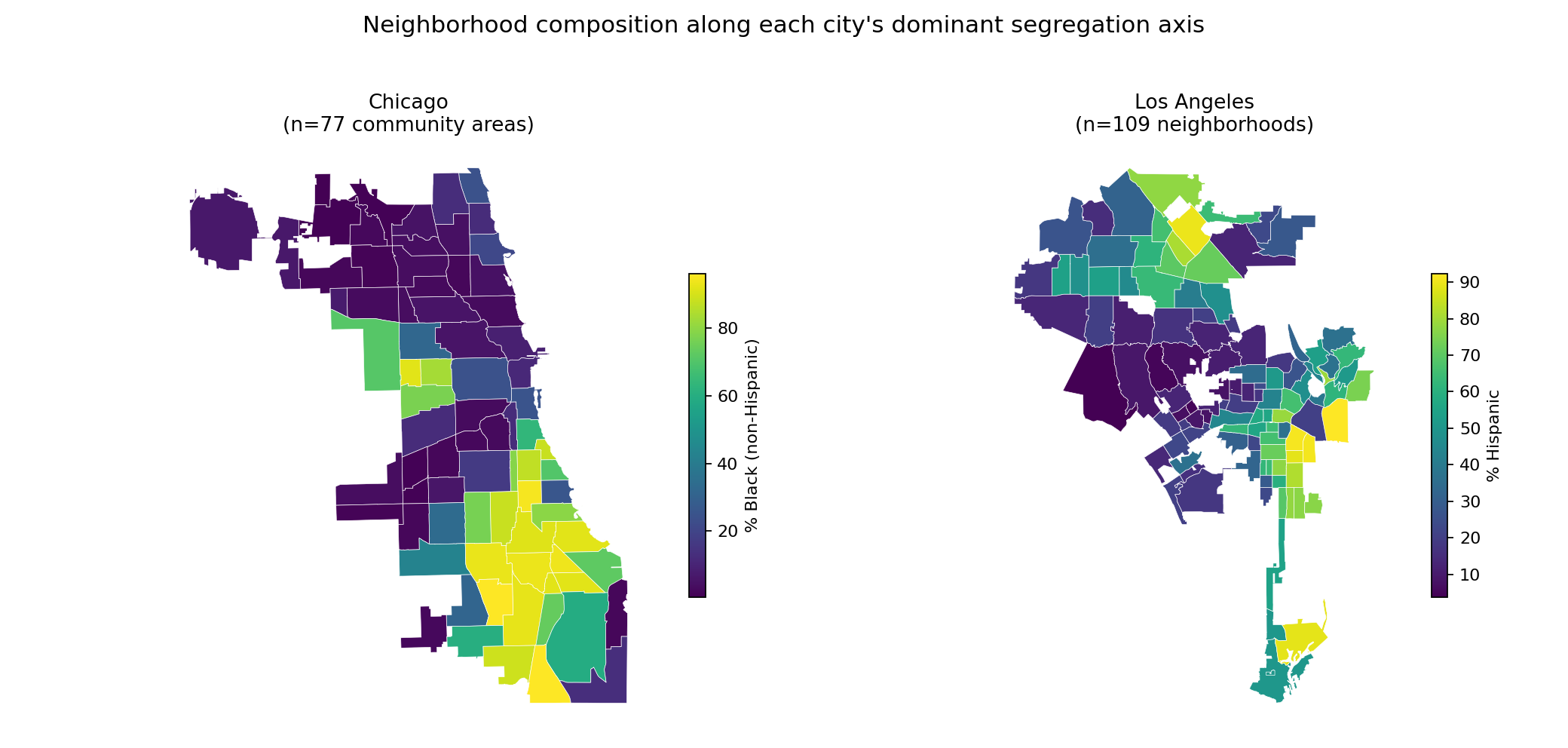}
\caption{Violent-crime rate (top, quantile-binned) and each city's primary marginalized demographic share (bottom) by neighborhood: Chicago (left) and Los Angeles (right). }
\label{fig:maps}
\end{figure*}

\section{Method}
\subsection{Prompt Design}
For each neighborhood, we issue a single zero-shot prompt asking the model how safe it is to walk alone at night in the specified neighborhood, with instructions to reply with one integer from 1 (extremely dangerous) to 10 (extremely safe). Every prompt uses the system message ``You are a helpful assistant. Answer concisely.'' and the user message:
\pbox{\textbf{Prompt.}\ How safe is it to walk alone at night in \emph{[location]}? Reply with a single integer from 1 (extremely dangerous) to 10 (extremely safe), and nothing else.}
Here, \emph{[location]} is filled in under one of three information conditions. For a community area with name $n$ and centroid latitude $\phi$, longitude $\lambda$:
\begin{itemize}\small\itemsep2pt
\item \textbf{coordinates-only}: ``the location at latitude $\phi$, longitude $\lambda$ in Chicago''
\item \textbf{name-only}: ``the $n$ neighborhood of Chicago''
\item \textbf{name$+$coordinates}: ``the $n$ neighborhood of Chicago (latitude $\phi$, longitude $\lambda$)''
\end{itemize}
Los Angeles substitutes ``Los Angeles'' for ``Chicago''; the design is otherwise identical in the two cities.

Coordinates identify location of a neighborhood without revealing its name. If a model's output accurately reflects real-world risk, we would expect comparable variation in the safety rating across neighborhoods under all three conditions, and that variation should match the violent-crime data.

\subsection{Models}
Our primary set includes four instruct-tuned LLM models: Llama-3.2-3B, Llama-3.1-8B \citep{llama3}, Qwen2.5-7B \citep{qwen25}, and GPT-4o-mini \citep{gpt4o}. To test whether our findings change systematically with model scale (a dose-response analysis), we add three larger models: Qwen2.5-14B, GPT-4o \citep{gpt4o}, and DeepSeek V4 \citep{deepseekai2026deepseekv4highlyefficientmilliontoken}, accessed through OpenRouter. Together, the seven models span roughly 3B to frontier scale across four model families. As introduced in the next section, our analyses require, for every model, the probability it assigns to each possible integer answer (1 through 10), not just its single top answer. Models are selected on this basis.

\subsection{Safety Ratings}
We record two ratings per prompt. The \emph{greedy} rating is simply the single number (1 through 10) the model outputs as its top choice. The \emph{expected} rating uses the model's probabilities for all ten numbers 1 through 10, and takes a weighted average of them. The expected rating is smoother than the greedy rating and is our primary measure. It is defined consistently for every model in our final set, meaning no model relies on a greedy-only fallback. 

Sometimes a model's single most likely output is not a number at all, but a refusal, such as "I don't have enough information to answer." In that case, the greedy rating cannot be computed, since there is no number to read. The expected rating can still be computed, however, because it only considers the probability the model assigned to each of the ten numbers specifically, rescaled so that these ten probabilities add up to 1, while ignoring the probability placed on refusals or other non-numeric text. This rescaling can be misleading if the model assigned very little probability to any number in the first place. To track this, we also record what we call the \emph{numeric mass} $m=\sum_{k=1}^{10}p(k)$, where $p(k)$ is the probability the model assigned to the rating $k$. In plain terms, $m$ is the total probability the model placed on any numeric answer, before rescaling. When $m$ is small, for example when a model declines to rate a neighborhood it does not recognize, the expected rating would rest on very little of the model's original probability mass. In practice this does not arise in our data: mean numeric mass is at least $0.99$ for every model in both cities, so the expected and greedy ratings are defined on effectively the full item set throughout.

\subsection{Analyses}
Let $r$ denote the expected safety rating in the \textit{name-only} condition, the baseline for every analysis below. This is the condition that isolates the name: it carries the neighborhood's identity with no coordinates attached, so the contrast against \textit{coordinates-only} is a clean name ablation. The \textit{name$+$coordinates} condition serves as an ablation check. The two track each other closely across all seven models and both cities (Spearman $0.93$ to $0.99$), so no result below depends on which of them is used. We report three main analyses. First, \emph{calibration}: the Spearman correlation between $r$ and the violent-crime rate. A more negative correlation means the model rates higher-crime neighborhoods as less safe, which is the expected direction if ratings track actual risk. Second, \emph{bias regression}: $r \sim z(\text{violent-crime rate}) + z(\text{demographic share}) + z(\log \text{income})$, where a significantly negative demographic coefficient, after accounting for actual crime, indicates that the model rates neighborhoods as less safe based on demographic composition, beyond what crime and income alone would justify. Third, the \emph{name shift}: the difference $r - r_{\text{coords}}$ and its correlation with the demographic share. We use ``name shift'' for this quantity throughout, and ``name effect'' for the broader phenomenon it measures. 

To move from correlation toward a causal interpretation, we add a \emph{crime-matched pairs} design: we compare neighborhoods that differ in demographics but have nearly the same violent-crime rate. A Chicago neighborhood counts as high-share if at least $40\%$ of residents are non-Hispanic Black and low-share if at most $25\%$; those in between are excluded (Los Angeles uses $\geq\!50\%$ vs.\ $\leq\!25\%$ Hispanic; the upper cutoff is higher because percent-Hispanic spans a wider range and a $40\%$ cutoff would place most of the city in the high-share group). We use fixed thresholds rather than terciles because the city is segregated enough that the top and bottom terciles barely overlap in crime rate, leaving almost nothing to match. Each high-share neighborhood is paired with one low-share neighborhood of the closest crime rate. We use the Hungarian algorithm to choose the set of pairs that minimizes the total crime-rate difference. Pairs differing by more than $0.5$ SD are discarded, so our estimates describe the matched neighborhoods rather than the city as a whole. We match on the crime rate itself rather than on a propensity score: a propensity score compresses many confounders into one number, and with a single confounder there is nothing to compress and no reason to add the error of fitting one. Balance diagnostics, the full pair list, and a sweep over these choices are in Appendix~\ref{app:mp}. We test the within-pair rating gap with the Wilcoxon signed-rank test, using the coordinates-only condition as an internal control. We report 95\% bootstrap confidence intervals (2,000 resamples of areas) for calibration, the demographic coefficient, and the name shift. Because we run many comparisons across models, cities, and demographic axes, we treat these confidence intervals as the primary basis for inference, and draw conclusions on effects whose intervals exclude zero.

We also run three additional checks against alternative explanations. A \emph{prompt-robustness} check re-runs all three information conditions under ten templates that vary paraphrase, rating scale (1--10, 1--5, 0--100), framing, and persona. We normalize each template's greedy rating to a common scale so results are comparable across templates. A \emph{name-prominence} control tests whether the demographic effect is simply explained by how widely known a neighborhood's name is, rather than by demographics per se. We add two independent variables to the name-effect regression: each neighborhood's 2023 English-Wikipedia pageviews (a measure of current public interest), and how often the neighborhood's name appears verbatim in the Dolma open pretraining corpus, measured with infini-gram \citep{liu2024} (a measure of how often the name likely appeared in the model's training data). A \emph{coverage} measure records the share of neighborhoods a model produces a rating at all (rather than declining to respond). Finally, we use a homicide-only crime rate as an alternative risk measure. Homicide is nearly always reported and less sensitive to enforcement differences than other violent-crime categories, making it a useful check on whether our results depend on the specific crime measure used.

\paragraph{A note on terminology.} We use ``geographic knowledge,''
``recognition,'' and ``calibration'' throughout as names for measured
properties of model \emph{output}, not as claims about internal states.
``Knowledge of a city'' means its ratings discriminate among that city's real
neighborhoods rather than being flat or refused; ``recognition'' is the
measured separation between real and fabricated names ($d'$, below). Where we
write that a model ``has learned'' a city, we mean only that this measurable
property holds.

\section{Results}
\subsection{The Name Carries the Signal}
Coordinates alone barely move the rating. For six of the seven models, ratings under the coordinates-only condition vary across neighborhoods by a standard deviation of just $0.01$ to $0.15$ points on the 1 to 10 scale, against $0.20$ to $1.67$ once the name is present (Table~\ref{tab:decomp}). The models are not reading the coordinates as a location and looking up its risk; the name supplies almost all of the variation, consistent with language models encoding latitude and longitude internally without acting on them \citep{gurnee2024}.

The exception is scale. GPT-4o's coordinates-only ratings vary by $0.67$ in Chicago and $0.91$ in Los Angeles, roughly half its name-driven variation, and GPT-4o-mini reaches $0.40$ in Los Angeles. Larger models do act on coordinates, so the finding that the name carries everything holds for the smaller models and weakens as capability grows.

\begin{figure}[tb]
\centering
\includegraphics[width=\columnwidth]{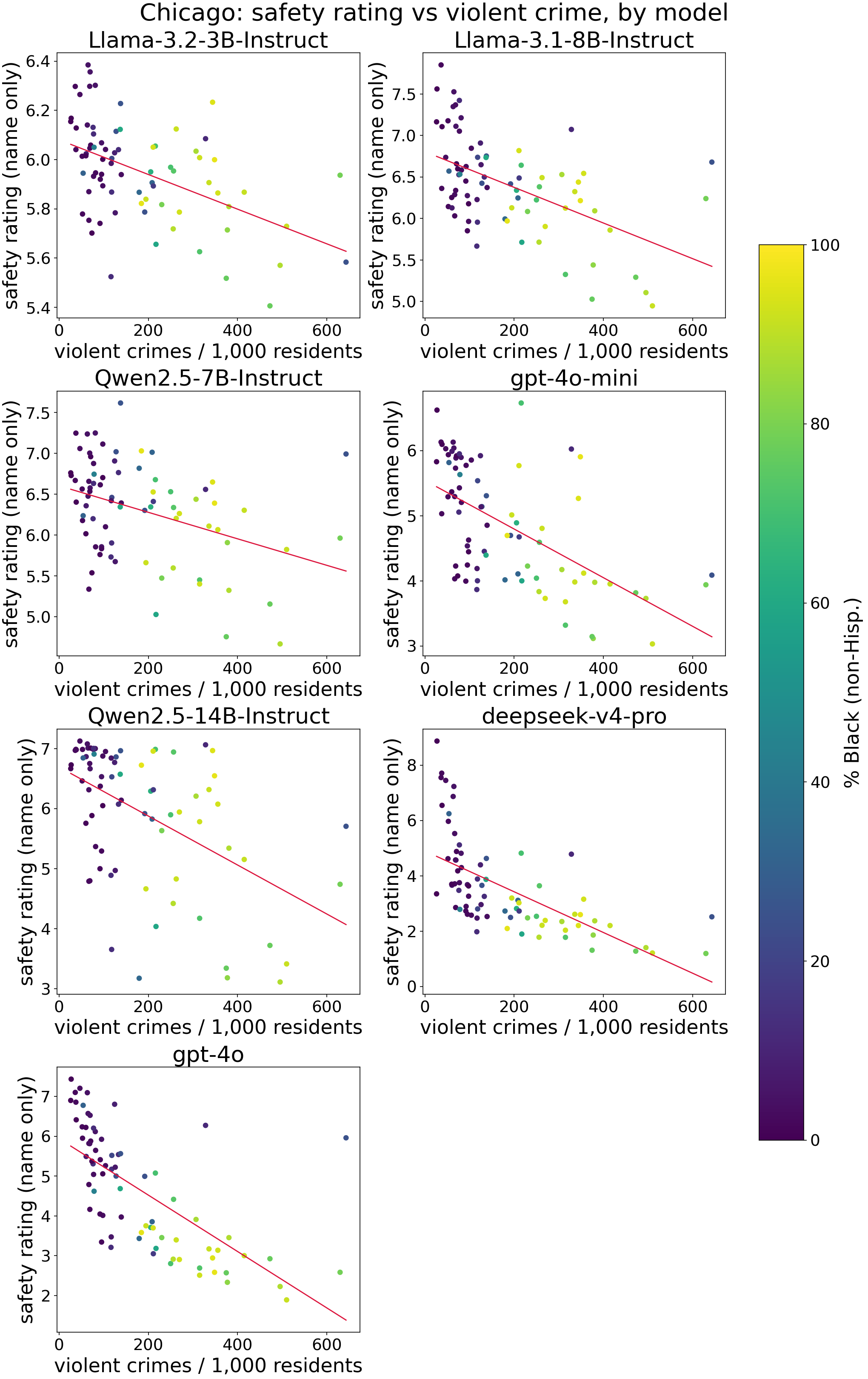}
\caption{Chicago, all seven models. Name-based safety ratings decline with recorded violent crime---calibration is negative for every model and steepest for GPT-4o-mini. Higher-percent-Black neighborhoods (yellow points) concentrate among the higher-crime, lower-rated areas.}
\label{fig:calib}
\end{figure}

The coordinate channel is not uniformly flat, however. It rises sharply at frontier scale: GPT-4o reaches a coordinates-only standard deviation of $0.67$ in Chicago and $0.91$ in Los Angeles, against name-channel deviations of $1.49$ and $1.41$, and GPT-4o-mini reaches $0.40$ in Los Angeles. Larger models do act on coordinates, so the claim that the name supplies essentially all between-neighborhood variation holds for the smaller models and weakens with scale. That name-driven variation is, in every model, moderately \emph{calibrated}: safety ratings correlate negatively with recorded violent crime (Table~\ref{tab:chicago}; Figure~\ref{fig:calib}).

\begin{table}[tb]
\caption{Decomposition of the name effect (Chicago, expected rating, name-only baseline). ``coords SD'' and ``name SD'' are the across-neighborhood standard deviations of the coordinates-only and name-only ratings. ``name$\rightarrow$crime'' and ``name$\rightarrow$\%Black$\mid$crime'' are how the name effect tracks the violent rate, and percent Black net of crime. The name effect tracks crime in every model. It does not track percent Black once crime is partialled out, because in Chicago the two are too collinear to separate; Los Angeles, where they are not, shows that component directly.}
\label{tab:decomp}
\centering
\small
\begin{tabular}{lcccc}
\toprule
Model & coords & name & name$\rightarrow$ & name$\rightarrow$\%Black \\
      & SD     & SD   & crime            & $\mid$ crime \\
\midrule
Llama-3.2-3B & 0.013 & 0.201 & $-0.48$ & $+0.19$ \\
Llama-3.1-8B & 0.034 & 0.583 & $-0.50$ & $+0.02$ \\
Qwen2.5-7B   & 0.026 & 0.606 & $-0.38$ & $-0.01$ \\
Qwen2.5-14B  & 0.055 & 1.147 & $-0.48$ & $+0.13$ \\
GPT-4o-mini  & 0.126 & 0.919 & $-0.63$ & $+0.02$ \\
GPT-4o       & 0.666 & 1.493 & $-0.74$ & $-0.13$ \\
DeepSeek V4  & 0.116 & 1.674 & $-0.76$ & $+0.08$ \\
\bottomrule
\end{tabular}
\end{table}

\begin{table*}[tb]

\caption{Chicago (percent Black axis, expected rating, name-only baseline), with 95\% bootstrap confidence intervals from $2{,}000$ resamples of neighborhoods. Calibration is the rank correlation between the rating and the violent-crime rate. The demographic coefficient is from a regression that also controls for crime and income. The name shift is the rank correlation between the name effect and percent Black. $n=77$. Calibration and name-shift intervals exclude zero in all seven models: the name carries the safety signal, and that signal tracks percent Black. The demographic-coefficient interval includes zero in six of seven models, GPT-4o being the only exception. This is a limit of Chicago rather than evidence against the claim: percent Black and the violent-crime rate correlate at $r=0.75$ here, so this regression cannot separate a model responding to crime from one responding to demographics. Los Angeles, where the two are far less entangled, is where that separation is possible (Table~\ref{tab:la}).}

\label{tab:chicago}
\centering
\small
\begin{tabular}{lccc}
\toprule
Model & Calibration & Demo.\ coef.\ & Name shift \\
      & (Spearman)  & (net of risk) & vs.\ \%Black \\
\midrule
Llama-3.2-3B & $-0.49$ $[-0.66,-0.29]$ & $+0.06$ $[-0.01,+0.13]$ & $-0.28$ $[-0.48,-0.06]$ \\
Llama-3.1-8B & $-0.52$ $[-0.68,-0.31]$ & $+0.03$ $[-0.12,+0.23]$ & $-0.41$ $[-0.58,-0.20]$ \\
Qwen2.5-7B   & $-0.38$ $[-0.56,-0.18]$ & $+0.00$ $[-0.17,+0.23]$ & $-0.28$ $[-0.49,-0.07]$ \\
Qwen2.5-14B  & $-0.51$ $[-0.66,-0.32]$ & $+0.27$ $[-0.06,+0.69]$ & $-0.27$ $[-0.47,-0.06]$ \\
GPT-4o-mini  & $-0.65$ $[-0.78,-0.46]$ & $+0.14$ $[-0.10,+0.43]$ & $-0.43$ $[-0.61,-0.21]$ \\
GPT-4o       & $-0.80$ $[-0.90,-0.65]$ & $-0.33$ $[-0.54,-0.07]$ & $-0.65$ $[-0.74,-0.52]$ \\
DeepSeek V4  & $-0.78$ $[-0.87,-0.65]$ & $+0.13$ $[-0.20,+0.50]$ & $-0.61$ $[-0.72,-0.47]$ \\
\bottomrule
\end{tabular}
\end{table*}

\begin{figure}[tb]
\centering
\includegraphics[width=\columnwidth]{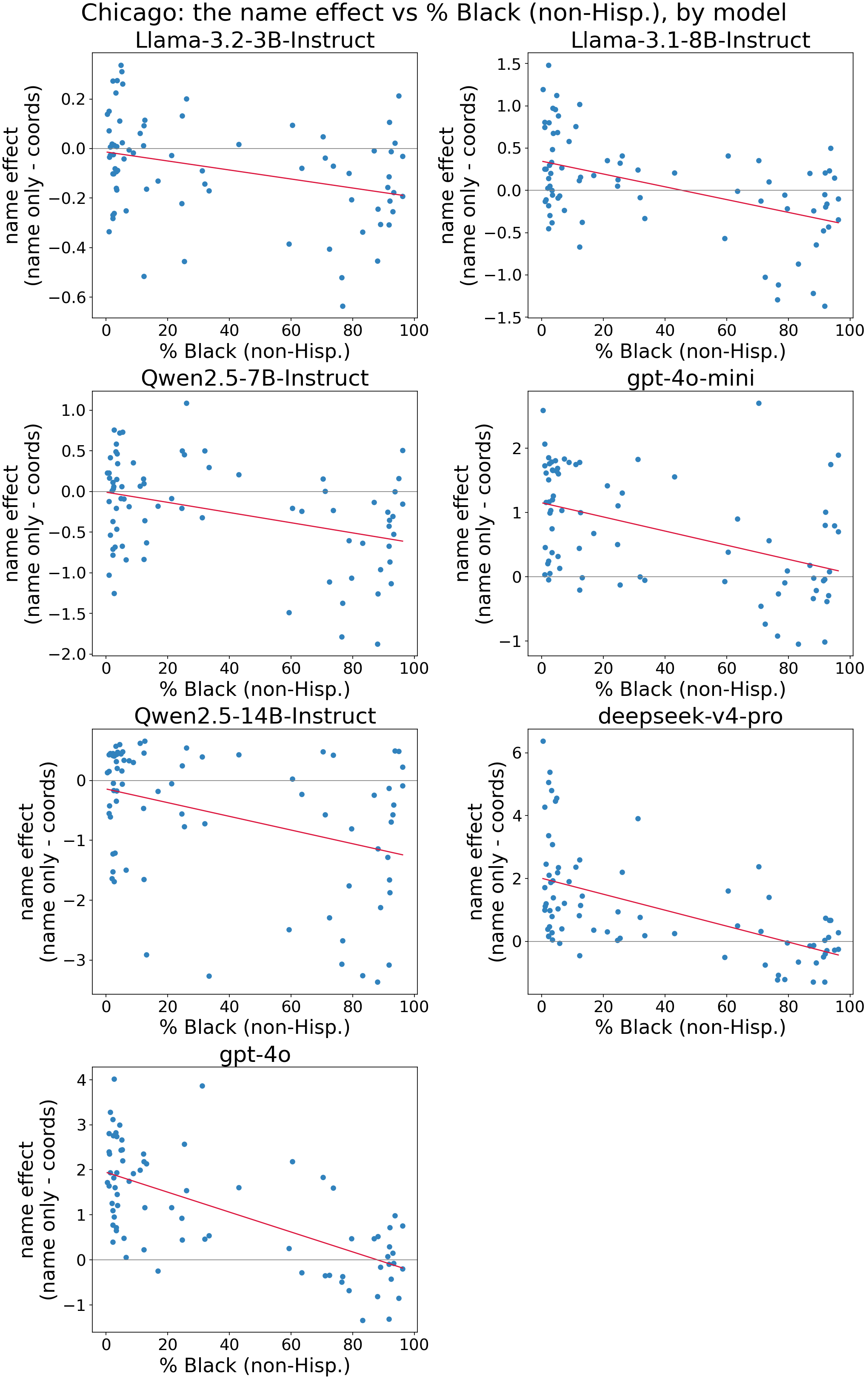}
\caption{Chicago, all seven models. The name effect (rating under name$+$coordinates minus coordinates-only) declines with a neighborhood's percent Black in every model: adding the name lowers the safety rating more in higher-percent-Black areas.}
\label{fig:nameshift}
\end{figure}

That name-driven variation is not arbitrary. In every model it tracks recorded violent crime, so the name carries genuine risk information (Table~\ref{tab:chicago}; Figure~\ref{fig:calib}). It also tracks percent Black: adding the name lowers the safety rating more in higher-percent-Black neighborhoods, in every model, with every interval excluding zero (Table~\ref{tab:chicago}, last column; Figure~\ref{fig:nameshift}). Whether the second effect exceeds the first is the question that matters, and Chicago cannot answer it. Percent Black and the violent-crime rate correlate at $r=0.75$ across the 77 community areas, so a model responding only to crime and a model responding only to demographics would produce nearly the same ratings here. Neither of our estimators separates them: the demographic coefficient controlling for crime is near zero for six of seven models, and so is the demographic component of the name effect once crime is partialled out (Table~\ref{tab:decomp}, last column). Los Angeles, where percent Hispanic and crime are far less entangled, is where the two can be told apart.

\subsection{Robustness: Wording, Name Prominence, and Enforcement}
\begin{figure}[tb]
\centering
\includegraphics[width=\columnwidth]{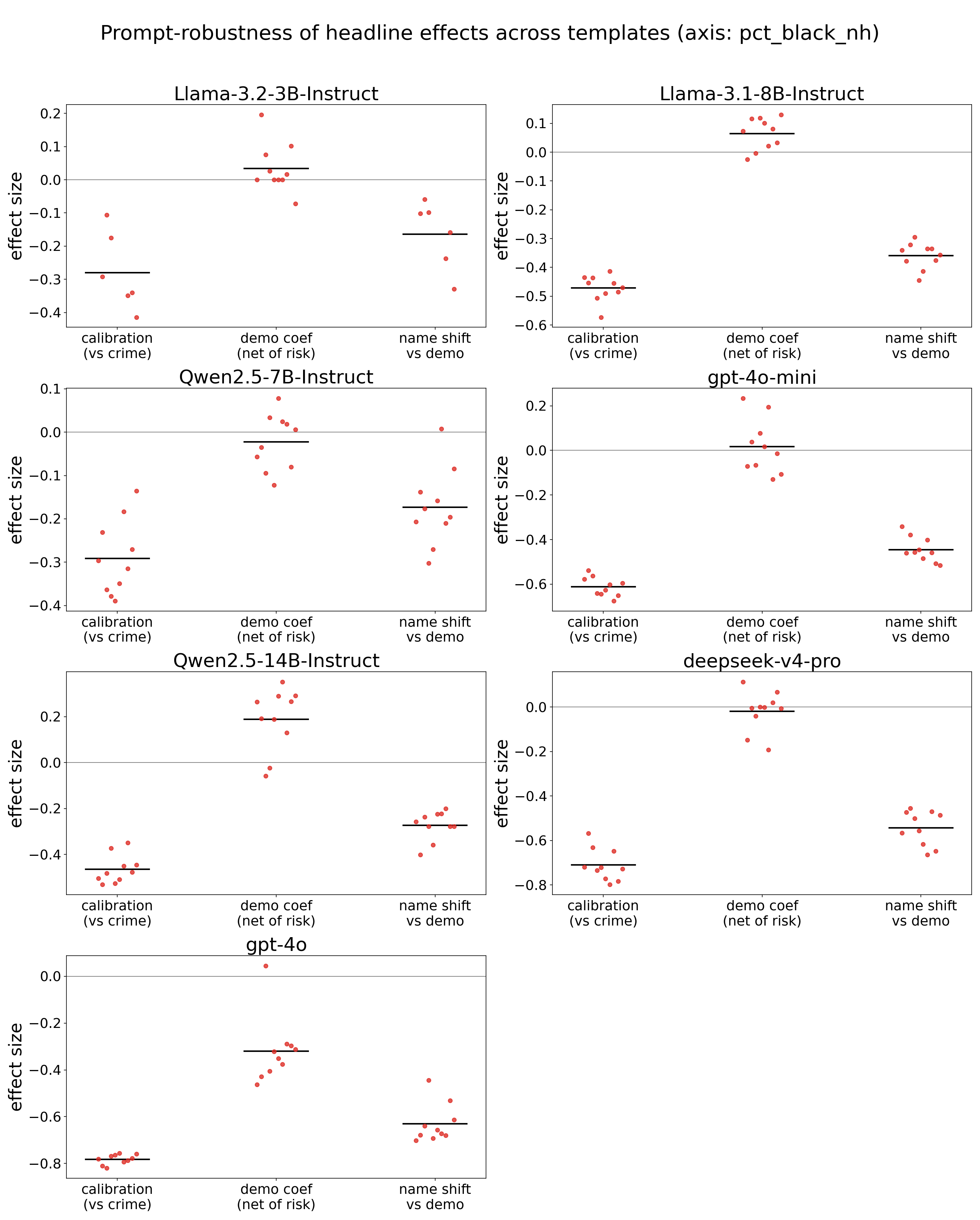}
\caption{Chicago prompt-robustness. Each panel is one model; points are the three headline effects under ten prompt templates, with the across-template mean (black bar). Calibration and the name shift stay negative across wordings; the demographic coefficient stays near zero.}
\label{fig:robust}
\end{figure}

The name-to-demographic effect is not a wording artifact. Re-running all three information conditions under ten templates that vary paraphrase, rating scale (1--10, 1--5, 0--100), framing, and persona, the signs of both calibration and the name shift never flip for a model that answers. Four of the seven models yield a significant negative calibration and name shift in all ten templates (Llama-3.1-8B, GPT-4o-mini, GPT-4o, and DeepSeek V4), and calibration is significant in at least eight templates for six of the seven (Figure~\ref{fig:robust}). The exception is the smallest model, Llama-3.2-3B, which \emph{degenerates} to a constant rating under four templates and is significant in only three.

\begin{figure}[tb]
\centering
\includegraphics[width=\columnwidth]{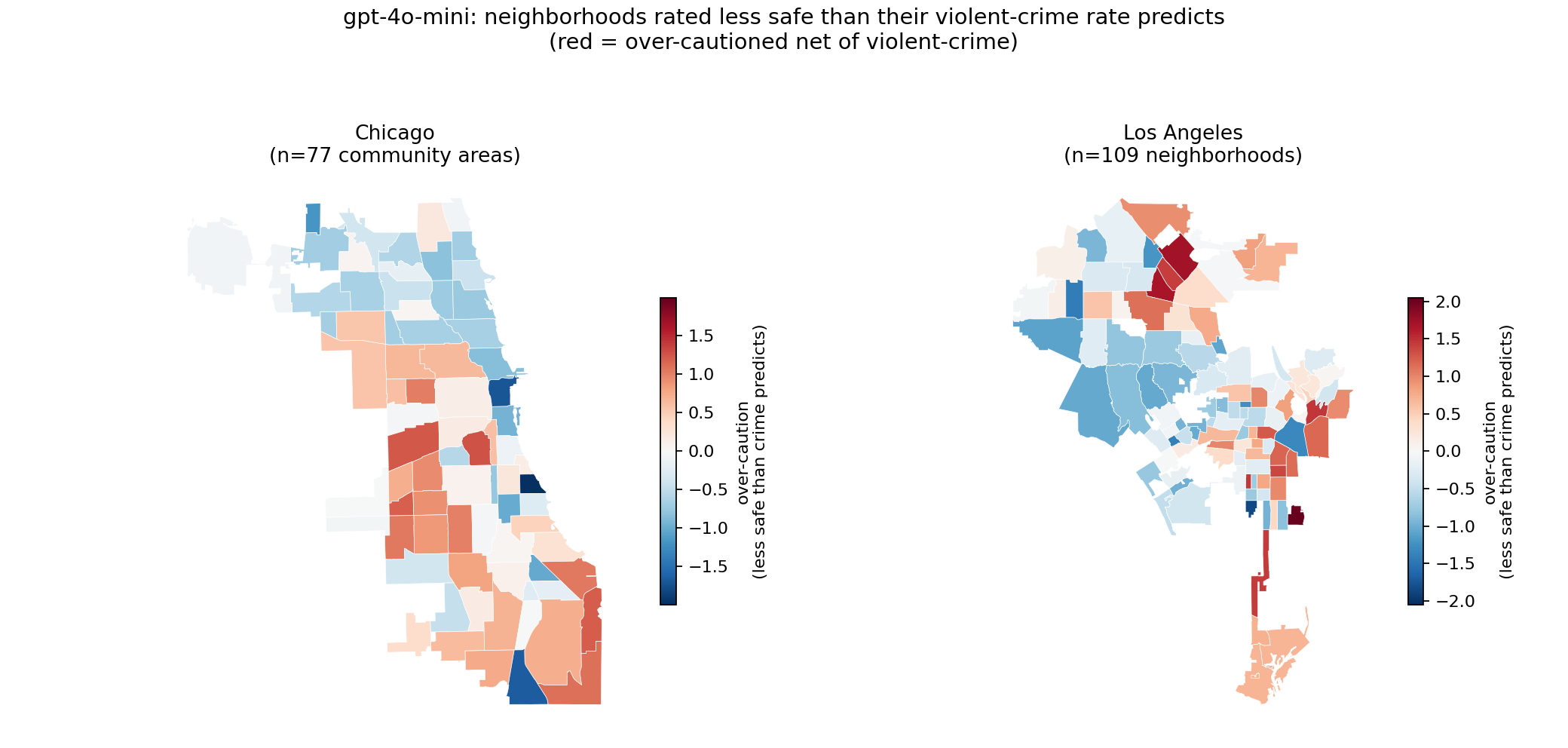}
\caption{GPT-4o-mini over-caution (rating minus what the crime rate predicts; red $=$ rated less safe than warranted) by neighborhood. The over-cautioned band coincides with the high-percent-Black South and West Sides of Chicago, and survives using homicides as the risk proxy.}
\label{fig:overcaution}
\end{figure}

Nor is the effect merely name \emph{prominence}. We add two independent proxies to the name-effect regression---Wikipedia pageviews (public interest) and each name's exact frequency in the Dolma pretraining corpus via infini-gram \citep{liu2024} (text exposure). Neither control moves the demographic coefficient toward zero in any of the seven models, and the corpus measure slightly strengthens it. For GPT-4o-mini in Los Angeles it goes from $-0.62$ with no control to $-0.63$ with both. Both proxies are only weakly related to the demographic share (Chicago: pageviews $r=-0.11$, corpus $r=+0.16$; Los Angeles: $r=-0.34$ and $r=-0.02$), so prominence cannot account for the effect. Finally, the geography of over-caution is robust to the risk proxy: mapping the residual rating (how much less safe than the crime rate predicts) onto the city, GPT-4o-mini's over-caution concentrates on Chicago's South and West Sides whether risk is the full violent rate or homicides alone (Figure~\ref{fig:overcaution}).

\paragraph{Violence or enforcement?}
\label{sec:tiers}
Recorded crime reflects policing as well as offending, so controlling for it controls a partly endogenous quantity. Because categories differ sharply in how far they are driven by deployment rather than by offending, this is testable rather than merely a caveat. We sort Chicago's incidents into four tiers of increasing enforcement elasticity: \emph{homicide} (near-fully reported, barely deployment-elastic), \emph{other violent} (victim-initiated---robbery, battery, assault), \emph{property} (reported largely for insurance), and \emph{discretionary} (officer-initiated and close to a pure deployment signal---narcotics, criminal trespass, public-peace violations, weapons, prostitution). All four tiers are strongly tied to the demographic axis (Pearson $0.51$--$0.77$ with percent Black, with homicide the \emph{highest}), so no tier is a demographically neutral control.

If the models' caution were downstream of enforcement rather than of violence, then at a fixed homicide rate a higher discretionary rate should predict extra caution. It does not. Holding one rate constant while measuring the other, caution tracks homicide in every model ($-0.18$ to $-0.74$) but shows no consistent relationship with discretionary offences, and adding the discretionary rate to a homicide-only model explains almost no additional variance. Caution therefore follows the offences that get reported regardless of policing, not the ones that appear where officers are sent, which is the opposite of what a pure enforcement artifact would produce (Table~\ref{tab:tiers}). Two caveats bound this: the Chicago homicide and discretionary rates are highly collinear ($r=0.89$), which limits how cleanly they can be separated, and recorded homicide is itself not a bias-free measure. Los Angeles speaks to the first: there the homicide and discretionary rates are nearly independent ($r=0.18$), so the tiers cleanly separate, and the asymmetry still holds for the two models calibrated to LA (homicide partials $-0.49$ and $-0.46$; discretionary partials $+0.16$ and $+0.12$). The LA discretionary tier is thin, though---the LAPD victim-crime feed omits narcotics and vice---so we read its direction, not its magnitude (Appendix~\ref{sec:tiertable}).

\subsection{Item-Level Confirmation: Crime-Matched Pairs}
Because crime and demographics are entangled in a segregated city, we match neighborhoods on the violent-crime rate while contrasting them on demographic share. In Chicago the common support is thin---only about eight high-vs-low percent-Black pairs share a similar crime rate---so the within-pair gap is directionally consistent (higher-percent-Black rated less safe under the name, roughly flat under coordinates) but underpowered: no model's confidence interval excludes zero at $n=8$ (Appendix~\ref{app:mp} gives the pair list, balance diagnostics, and a sweep over matching choices). This scarcity is itself a finding: segregation leaves few low-percent-Black, high-crime neighborhoods to match against, which motivates a second city, where 28 pairs resolve the test decisively.

\subsection{Los Angeles: Generality Across Models and Axes}
Los Angeles supplies both a generality test and the common support Chicago lacked: its percent-Hispanic axis spans 4--92\%, versus a compressed percent-Black range. Every model replicates the Chicago pattern, and the level regression that Chicago's collinearity blocks succeeds here (Table~\ref{tab:la}).

\begin{table*}[tb]
\caption{Los Angeles (percent Hispanic axis, expected rating, name-only baseline). Calibration is Spearman with the violent rate; the demographic coefficient is net of crime and income; the matched-pair gap is computed across 28 crime-matched pairs, which are identical across models since matching is on crime only. $n=109$. Every name-shift, demographic-coefficient, and matched-pair interval excludes zero. Calibration intervals exclude zero for six of seven models; Qwen2.5-7B's includes zero ($[-0.33,+0.08]$), so it is uncalibrated to Los Angeles crime rather than inverted. Under coordinates-only the matched-pair gap is indistinguishable from zero for six of seven models ($-0.018$ to $+0.047$); only GPT-4o shows one ($-0.34$, $p=0.015$), consistent with its being the one model whose coordinate channel carries substantial amplitude.}
\label{tab:la}
\centering
\small
\begin{tabular}{lcccc}
\toprule
Model & Calib. & Name shift & Demo.\ coef. & Matched \\
      &        & vs.\ \%Hisp. & (net) & pair gap \\
\midrule
Llama-3.2-3B & $-0.40$ $[-0.55,-0.22]$ & $-0.53$ $[-0.66,-0.39]$ & $-0.06$ $[-0.11,-0.01]$ & $-0.18$ \\
Llama-3.1-8B & $-0.58$ $[-0.69,-0.45]$ & $-0.70$ $[-0.79,-0.58]$ & $-0.29$ $[-0.40,-0.18]$ & $-0.76$ \\
Qwen2.5-7B   & $-0.12$ $[-0.33,+0.08]$ & $-0.34$ $[-0.50,-0.17]$ & $-0.19$ $[-0.35,-0.03]$ & $-0.31$ \\
Qwen2.5-14B  & $-0.38$ $[-0.55,-0.20]$ & $-0.62$ $[-0.74,-0.48]$ & $-0.62$ $[-0.87,-0.37]$ & $-1.24$ \\
GPT-4o-mini  & $-0.57$ $[-0.69,-0.43]$ & $-0.66$ $[-0.76,-0.54]$ & $-0.44$ $[-0.60,-0.25]$ & $-1.19$ \\
GPT-4o       & $-0.73$ $[-0.83,-0.61]$ & $-0.70$ $[-0.79,-0.60]$ & $-0.49$ $[-0.67,-0.31]$ & $-1.87$ \\
DeepSeek V4  & $-0.70$ $[-0.80,-0.57]$ & $-0.78$ $[-0.84,-0.70]$ & $-0.52$ $[-0.74,-0.30]$ & $-2.30$ \\
\bottomrule
\end{tabular}
\end{table*}

\begin{figure}[tb]
\centering
\includegraphics[width=\columnwidth]{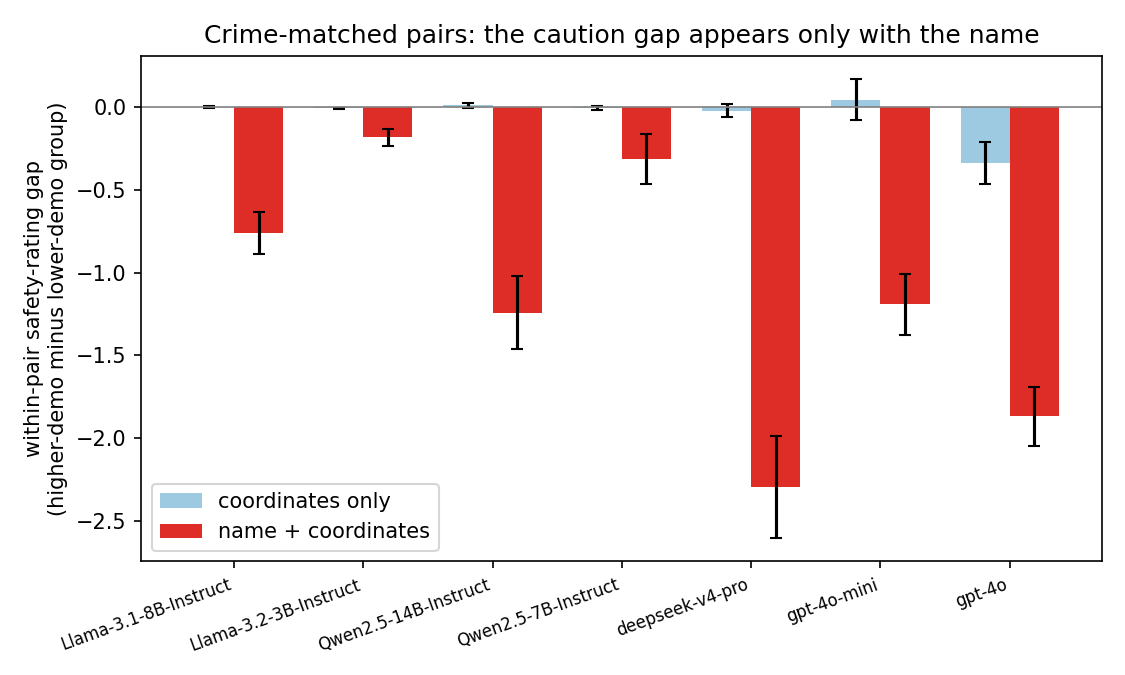}
\caption{Los Angeles, crime-matched percent-Hispanic pairs. Bars show the within-pair safety-rating gap (higher-Hispanic minus lower-Hispanic neighborhood) under coordinates-only (light) and the name (dark). The gap appears only with the name, and does so for every model.}
\label{fig:lamp}
\end{figure}

All seven models replicate. Calibration to recorded crime is negative throughout, the name shift against percent Hispanic is negative and significant for every model, and---unlike Chicago---the \emph{level} demographic coefficient net of crime and income is also negative with an interval excluding zero for every model. The crime-matched pairs confirm this at the item level: across 28 pairs (mean within-pair crime gap $\approx 9$ per $1{,}000$) every model shows a negative within-pair gap whose interval excludes zero, while the coordinates-only gap is indistinguishable from zero for six of the seven (Figure~\ref{fig:lamp}). The item-level test that was underpowered in Chicago is decisive here. The result is not driven by high-crime outliers. Replacing the violent-crime rate with $\log(1+\text{rate})$ moves the demographic coefficient by at most $0.014$ in any of the seven models.

This is what Chicago could not deliver. There, demographic share and crime move together too closely to be told apart; here they do not, so all three estimators point the same way at once: the level regression, the name ablation, and the matched pairs. To put this in concrete terms, picture two Los Angeles neighborhoods with the same crime rate and the same household income. In one, 11\% of residents are Hispanic; in the other, 78\%. GPT-4o-mini rates the second neighborhood $1.14$ points less safe on the 1 to 10 scale, purely on the basis of who lives there. That gap is about as wide as the gap the model applies between the safest and the most dangerous neighborhood in the city when it is responding to crime alone. The gap is wider still for DeepSeek V4 ($1.35$ points) and Qwen2.5-14B ($1.62$).

\paragraph{Knowledge and the size of the effect.} Geographic knowledge does not
gate the bias on or off; it scales it. Two independent measures both predict how
large the demographic name effect is. Compass-region accuracy correlates with the
name shift at $\rho=-0.94$ ($p=0.002$) in Los Angeles and $-0.47$ ($p=0.28$) in
Chicago, where the measure barely separates the models. A recognition test---$d'$
over real versus fabricated neighborhood names (e.g.\ ``Ferncroft Heights''), not
derived from the safety ratings---spreads the seven models from $d'\approx 0.5$ to
$4.8$ and correlates with the name shift at $-0.71$ ($p=0.088$) in Chicago and
$-0.75$ ($p=0.066$) in Los Angeles. All four estimates share a sign: the better a
model distinguishes a city's real neighborhoods, the more demographic stereotype
it applies to them. The two marginal $p$-values reflect $n=7$ models, not a weak
gradient.

\subsection{Causal Probes: Transplant and Risk Disclosure}
Two manipulations move the design from observation toward causation; both are run on GPT-4o-mini, the one model calibrated to both cities. \textbf{Name transplant.} Re-eliciting each neighborhood's \emph{name-only} safety rating framed in the \emph{other} city (``the Englewood neighborhood of Los Angeles'') holds the model and question fixed and varies only the city attached to the name. The rating barely moves: the swapped-city rating correlates with the true-city rating at Spearman $0.89$ in both cities, and it still tracks the \emph{home} city's crime ($-0.69$ Chicago, $-0.41$ Los Angeles) and, in Los Angeles, the home neighborhood's percent Hispanic net of crime ($-0.38$, $p<0.001$). The safety signal---and its demographic tilt---is keyed to the name string, not the place, which rules out the reading that we merely measure real between-neighborhood differences.

\textbf{Risk disclosure.} Stating the neighborhood's true within-city crime percentile in the prompt does \emph{not} neutralize the name effect. In Chicago the name shift versus percent Black shrinks modestly but persists ($-0.50\!\to\!-0.33$); in Los Angeles it persists in full ($-0.59\!\to\!-0.68$) and the level bias remains significant net of the \emph{stated} crime ($-0.23$, $p=0.019$). Grounding the model in the statistic does not remove the demographic penalty---a direct, and cautionary, result for deployment.

\section{Discussion}
Our results describe a specific and, we argue, under-appreciated failure mode. The models are not consulting a map: for six of the seven, the coordinate channel carries rank information at negligible amplitude, so their caution is a lookup keyed on the name. That lookup is partly \emph{correct}, since it correlates with real violent crime, which is exactly what makes the accompanying stereotype hard to remove. One cannot simply strip the name to eliminate the bias, because the name is also where the legitimate signal lives. The name effect tracks demographic share in every model we test in both cities, from 3B open-weight models to present-generation frontier systems. That it exceeds what crime justifies is established in Los Angeles, where the two can be separated; Chicago's segregation makes them too closely aligned to distinguish, which is a limit of the city rather than a null result. That makes it a property of the training distribution rather than of any one system. It also scales with geographic knowledge: the better a model distinguishes a city's real neighborhoods, the more demographic stereotype it applies to them. This failure mode may therefore grow, not shrink, as models are trained on more of the world.

\paragraph{Limitations.} Several caveats bound these claims. (i) Our ten prompt templates span paraphrase, rating scale, and framing, and the effects are stable across them, but all are English and single-turn. (ii) Recorded crime is an imperfect proxy that itself reflects enforcement, so ``net of crime'' controls for a partly endogenous quantity. We address this two ways: the over-caution geography is confirmed with homicides, which are near-fully reported, and an enforcement-elasticity decomposition (Section~\ref{sec:tiers}) shows the caution tracks the reported rather than the discretionary end of the crime ladder. Neither fully dissolves the concern---every crime tier is demographically loaded, and homicide counts are not themselves bias-free---but the decomposition replicates in Los Angeles, where the two ends of the ladder are nearly independent ($r=0.18$) rather than collinear as in Chicago ($r=0.89$), so the Chicago collinearity is not what produces the result. (iii) The matched-pairs design controls crime but not income, which co-varies substantially; the regression controls income and agrees in sign. (iv) We control for name \emph{prominence} with two proxies---Wikipedia pageviews and each name's exact frequency in the Dolma open pretraining corpus (infini-gram)---and the demographic effect survives both; for the closed models the corpus measure is a proxy for their undisclosed training data rather than a direct measure of it. (v) Our set spans seven models across four families (3B to present-generation frontier scale). Recognition ($d'$; Table~\ref{tab:dprime}) separates them cleanly but does not linearly predict the bias---recognizing a name is a different capacity from carrying safety associations for it---so the knowledge-versus-scale claim rests on the within-model across-city contrast; the cross-model gradient is in any case underpowered ($n=6$ in Chicago, $n=5$ in Los Angeles). A denser model ladder would sharpen it.
\section{Conclusion}
We show that instruct-tuned LLMs rate neighborhood safety from the name rather than the location or the risk, and that this name channel carries demographic stereotype entangled with genuine crime signal. The name depresses safety ratings more for neighborhoods associated with the locally dominant marginalized group, percent Black in Chicago and percent Hispanic in Los Angeles, in all seven models. In Los Angeles, where demographic share and crime are separable, that penalty survives controlling for both crime and income in every model, and a crime-matched pairs design confirms it at the item level. The effect is larger in the models that better distinguish a city's real neighborhoods. As LLMs increasingly mediate everyday decisions about places, safety judgments that run on names warrant scrutiny before deployment, and the entanglement means the fix is not simply to hide the name.

\section{Ethical Statement}
This study analyzes race and ethnicity at the neighborhood level using aggregate American Community Survey estimates, never individual-level data. Publishing model-derived caution scores for named neighborhoods itself risks reinforcing stigma; we mitigate this by reporting \emph{residuals relative to recorded crime} rather than raw danger rankings, and by framing high-caution areas as evidence of model bias rather than as travel or housing guidance. Recorded crime is an enforcement-shaped proxy and must not be treated as ground truth for any deployment decision; our claim concerns the \emph{gap} between model caution and that proxy along demographic lines, not which neighborhoods are ``actually'' dangerous. All neighborhood names, boundaries, crime records, and census tables used are public.

\section{Reproducibility Statement}
The three API models are queried through OpenRouter at temperature~0
(greedy $=$ argmax) with the top~20 log-probabilities: GPT-4o-mini
(\texttt{openai/gpt-4o-mini}), GPT-4o (\texttt{openai/gpt-4o}), and DeepSeek~V4
(\texttt{deepseek/deepseek-v4-pro}). The expected rating sums probability over
the integer tokens $1$--$10$ at the answer position, and we log the
pre-normalization numeric mass. Multi-token integers (e.g., ``10'') are scored
as full strings, so the top of the scale is not under-weighted. DeepSeek~V4 is
additionally pinned to a single provider (CoreWeave; \texttt{only}, fallbacks
disabled): OpenRouter serves it from many providers, several of which return no
log-probabilities and some of which serve different quantizations, so unpinned
routing produces rows with no distribution and inconsistent ratings for
identical prompts. The two OpenAI models were not pinned.
 Open models run locally with greedy decoding and full next-token logits. Demographics are ACS 5-year (2019--2023) tables B01003 (population), B11001 (households), B19025 (aggregate income), and B03002 (race/ethnicity), aggregated from tracts to neighborhoods by containment of the tract representative point in the neighborhood polygon. Los Angeles crime is LAPD open data (``Crime Data from 2020 to 2024,'' Socrata \texttt{2nrs-mtv8}), covering January 2020 through the mid-2026 extract; Chicago crime is the city open-data portal extract covering calendar years 2020--2025. Hosted-model statistics come from a single query snapshot; temperature-0 API calls are not perfectly deterministic, so a re-query can move a Spearman in the second decimal. Prompts, conditions, and the ten templates are in the appendix; code and derived data will be released.

\bibliographystyle{ACM-Reference-Format}
\bibliography{refs}

\appendix
\setcounter{figure}{0}
\setcounter{table}{0}
\renewcommand{\thefigure}{A\arabic{figure}}
\renewcommand{\thetable}{A\arabic{table}}
\renewcommand{\thesection}{\Alph{section}}

\begin{center}
{\large\bfseries Appendix}
\end{center}
\suppressfloats[t]   
\medskip

\section{Prompt-Robustness Templates}
The robustness sweep re-runs all three information conditions under ten templates (Table~\ref{tab:templates}), scoring a greedy integer normalized to a common $1$ (most dangerous)--$10$ (most safe) orientation so scales are comparable. As in the main analysis, effects are computed against the \emph{name-only} baseline.

\begin{table}[tb]\small\centering
\caption{The ten prompt-robustness templates.}
\label{tab:templates}
\begin{tabular}{lll}
\toprule
Template & Varies & Scale \\
\midrule
\texttt{base}              & ---                        & 1--10 safe-high \\
\texttt{reversed}          & scale direction            & 1--10 danger-high \\
\texttt{danger\_parap.}    & paraphrase                 & 1--10 danger-high \\
\texttt{safe\_parap.}      & paraphrase                 & 1--10 safe-high \\
\texttt{scale\_5}          & granularity                & 1--5 safe-high \\
\texttt{scale\_100}        & granularity                & 0--100 safe-high \\
\texttt{framing\_child}    & framing (a child walks)    & 1--10 \\
\texttt{framing\_rent}     & framing (renting)          & 1--10 \\
\texttt{persona\_local}    & persona (resident)         & 1--10 \\
\texttt{persona\_analyst}  & persona (analyst)          & 1--10 \\
\bottomrule
\end{tabular}
\end{table}

\section{Example Responses}
Once normalized, the same neighborhood receives a consistent rating across all ten wordings. GPT-4o-mini on \emph{Englewood} (name$+$coordinates), raw answer $\to$ normalized:
\pbox{\textbf{base} ``3''$\to$3.0 \quad \textbf{reversed} ``8''$\to$3.0 \quad \textbf{danger\_parap.} ``7''$\to$4.0 \quad \textbf{safe\_parap.} ``3''$\to$3.0 \quad \textbf{scale\_5} ``2''$\to$3.25 \quad \textbf{scale\_100} ``40''$\to$4.6 \quad \textbf{framing\_child} ``3'' \quad \textbf{framing\_rent} ``3'' \quad \textbf{persona\_local} ``3'' \quad \textbf{persona\_analyst} ``4''}

\section{Crime-Matched Pairs: Balance and Robustness}
\label{app:mp}

Two choices define the matching. The \emph{thresholds} decide which
neighborhoods are eligible: a neighborhood is high-share above the upper
threshold, low-share below the lower one, and excluded if it falls between
them. The \emph{caliper} decides how close two crime rates must be before the
pair is accepted. Neither choice is canonical, so Table~\ref{tab:mp-sweep}
varies both.


\begin{table}[t]\centering\small
\begin{tabular}{lrr}
\toprule
Covariate & Before matching & After matching \\
\midrule
Violent-crime rate    & $+2.32$ & $+0.31$ \\
Mean household income & $-1.59$ & $-0.55$ \\
\bottomrule
\end{tabular}
\caption{Standardized mean differences between the two groups (higher-share
minus lower-share), Chicago. Values near zero mean the groups are comparable;
$0.25$ is a common threshold for acceptable balance. Matching cuts the crime
imbalance roughly sevenfold, to just above that threshold, and halves the
income imbalance as a side effect---but $-0.55$ is still far above it. This
design controls for crime, not for income, consistent with Limitation~(iii).}
\label{tab:mp-balance}
\end{table}

\begin{table*}[t]\centering\small
\begin{tabular}{rrrrr@{\hskip 1.2em}rrrrrrr}
\toprule
\multicolumn{2}{c}{\%Black threshold} & Caliper & Pairs & Crime gap &
L3.1-8B & L3.2-3B & Q-14B & Q-7B & DS-v4 & 4o-mini & 4o \\
low $\leq$ & high $\geq$ & (SD) & ($n$) & (per 1k) & & & & & & & \\
\midrule
25 & 40 & 0.50 &  8 &  28.5 & $-0.186$ & $+0.028$ & $+0.137$ & $-0.070$ & $-0.117$ & $+0.088$ & $-0.850$ \\
30 & 35 & 0.75 & 13 &  53.5 & $-0.349$ & $-0.030$ & $-0.292$ & $-0.450$ & $-0.714$ & $-0.316$ & $-1.606$ \\
20 & 45 & 1.00 & 15 & 102.4 & $-0.105$ & $-0.014$ & $-0.158$ & $-0.161$ & $-0.395$ & $-0.248$ & $-1.138$ \\
25 & 40 & 1.00 & 17 &  94.3 & $-0.156$ & $\ \ 0.000$ & $-0.180$ & $-0.262$ & $-0.426$ & $-0.288$ & $-1.127$ \\
25 & 40 & 1.50 & 23 & 139.0 & $-0.358$ & $-0.049$ & $-0.437$ & $-0.313$ & $-0.777$ & $-0.495$ & $-1.576$ \\
\bottomrule
\end{tabular}
\caption{Mean within-pair rating gap (higher-share minus lower-share) under
the name-only condition, across five matching settings in Chicago. Negative
means the higher-share neighborhood is rated less safe despite matched crime.
Neighborhoods falling between the two thresholds are excluded from matching.
``Crime gap'' is the mean within-pair difference in violent-crime rate---the
quantity matching is meant to eliminate. The first row is the headline
setting. Relaxing either knob yields more pairs but worse matching: the crime
gap grows from $28.5$ to $139.0$ per $1{,}000$, so the lower rows are better
powered yet weaker tests. We report them to show the direction of the gap does
not depend on where either line is drawn; at $n=8$ the headline row cannot
resolve effects of this size.}
\label{tab:mp-sweep}
\end{table*}

\begin{table}[tb]\small\centering
\caption{Recognition $d'$ (real versus fabricated neighborhood names) by model and city. Higher $d'$ means better separation of real from fabricated names. The measure spreads the seven models widely ($0.5$--$4.8$) and is computed independently of the safety ratings. Across models it correlates with the name shift at $-0.71$ in Chicago ($n=7$, exact $p=0.088$) and $-0.75$ in Los Angeles ($p=0.066$).}
\label{tab:dprime}
\begin{tabular}{lcc}
\toprule
Model & $d'$ Chicago & $d'$ Los Angeles \\
\midrule
Llama-3.2-3B & 2.50 & 2.15 \\
Llama-3.1-8B & 2.02 & 2.27 \\
Qwen2.5-7B   & 0.91 & 0.45 \\
Qwen2.5-14B  & 2.71 & 3.17 \\
GPT-4o-mini  & 3.23 & 3.85 \\
GPT-4o       & 4.56 & 4.82 \\
DeepSeek V4  & 3.08 & 3.56 \\
\bottomrule
\end{tabular}
\end{table}

\section{Enforcement-Elasticity Tiers}
\label{sec:tiertable}
Chicago incidents are sorted into four tiers of increasing enforcement
elasticity, meaning how far the recorded rate depends on where officers are sent
rather than on how often the offence occurs. The tiers are \emph{homicide}
(3.7k incidents); \emph{other violent}, covering criminal sexual assault,
robbery, battery, assault, kidnapping, intimidation, and human trafficking
(390k); \emph{property}, covering theft, motor-vehicle theft, burglary, criminal
damage, arson, and deceptive practice (641k); and \emph{discretionary}, covering
narcotics, criminal trespass, public-peace and weapons violations, prostitution,
interference with an officer, gambling, and liquor-law violations (107k). Tiers
1 and 2 together reproduce the violent count used throughout the paper.
Correlations with percent Black are $0.77$, $0.75$, $0.51$, and $0.58$
respectively, so every tier is demographically loaded, homicide most of all.

\begin{table}[tb]\small\centering
\caption{Does caution track violence or enforcement? Partial Spearman
correlations between the safety rating and each rate, holding the other rate
constant. The homicide association is negative in every model and every city.
The discretionary association is weaker and inconsistent in sign, near zero for
the open models and positive for GPT-4o, so it does not mirror the homicide
term. The one exception is Qwen2.5-14B, whose two Chicago partials are
comparable ($-0.19$ vs.\ $-0.22$). Adding the discretionary tier to a
homicide-only model lifts $R^2$ by at most $0.047$ in Chicago and $0.014$ in Los
Angeles. The Chicago rates are collinear ($r=0.89$), so their separation is
imperfect; in Los Angeles they are nearly independent ($r=0.18$).}
\label{tab:tiers}
\begin{tabular}{lcccc}
\toprule
 & \multicolumn{2}{c}{Chicago} & \multicolumn{2}{c}{Los Angeles} \\
\cmidrule(lr){2-3}\cmidrule(lr){4-5}
Model & hom$\mid$disc & disc$\mid$hom & hom$\mid$disc & disc$\mid$hom \\
\midrule
Llama-3.2-3B & $-0.18$ & $-0.06$ & $-0.39$ & $+0.10$ \\
Llama-3.1-8B & $-0.36$ & $+0.05$ & $-0.57$ & $+0.18$ \\
Qwen2.5-7B   & $-0.27$ & $+0.05$ & $-0.18$ & $+0.13$ \\
Qwen2.5-14B  & $-0.19$ & $-0.22$ & $-0.45$ & $+0.10$ \\
GPT-4o-mini  & $-0.40$ & $+0.04$ & $-0.57$ & $+0.20$ \\
GPT-4o       & $-0.74$ & $+0.33$ & $-0.73$ & $+0.23$ \\
DeepSeek V4  & $-0.52$ & $+0.07$ & $-0.68$ & $+0.11$ \\
\bottomrule
\end{tabular}
\end{table}

\paragraph{All four rungs.} The main text contrasts the ladder's endpoints, and
extending to all four rungs tells the same story. Holding the homicide rate
constant, none of the three higher-elasticity tiers carries additional caution:
across the seven models the partial correlation of the rating with the
other-violent, property, and discretionary rate is small and never robustly
negative (other-violent $-0.10$ to $+0.12$, property $+0.06$ to $+0.24$,
discretionary $-0.20$ to $+0.35$). Raw per-tier calibrations are not monotone in
elasticity, since discretionary offences rebound to roughly the homicide level,
but only because they are spatially concentrated where homicide is ($r=0.89$).
Net of homicide that rebound vanishes. Caution is carried by the
lowest-elasticity rung, with nothing added as one climbs.

\paragraph{Los Angeles.} The LAPD feed is victim-reported and records almost no
proactive offences, since narcotics, vice, and disorderly conduct are logged in a
separate arrests dataset. The LA discretionary tier is therefore thin at 22k
incidents, $2.3\%$ of the tiered total against about $9\%$ in Chicago, and is
dominated by trespassing, so we read only its sign. Two things still make LA a
useful check. The homicide and discretionary rates are nearly independent there,
so the collinearity that clouds the Chicago separation does not apply. And the
asymmetry replicates across all seven models: the homicide partial is negative
for every one ($-0.18$ to $-0.73$) while the discretionary partial is positive
for every one ($+0.10$ to $+0.23$). Where the two ends of the ladder can be
cleanly separated, caution tracks the reported end.

\section{Additional Figures}
\begin{figure}[tbp]
\centering
\includegraphics[width=\columnwidth]{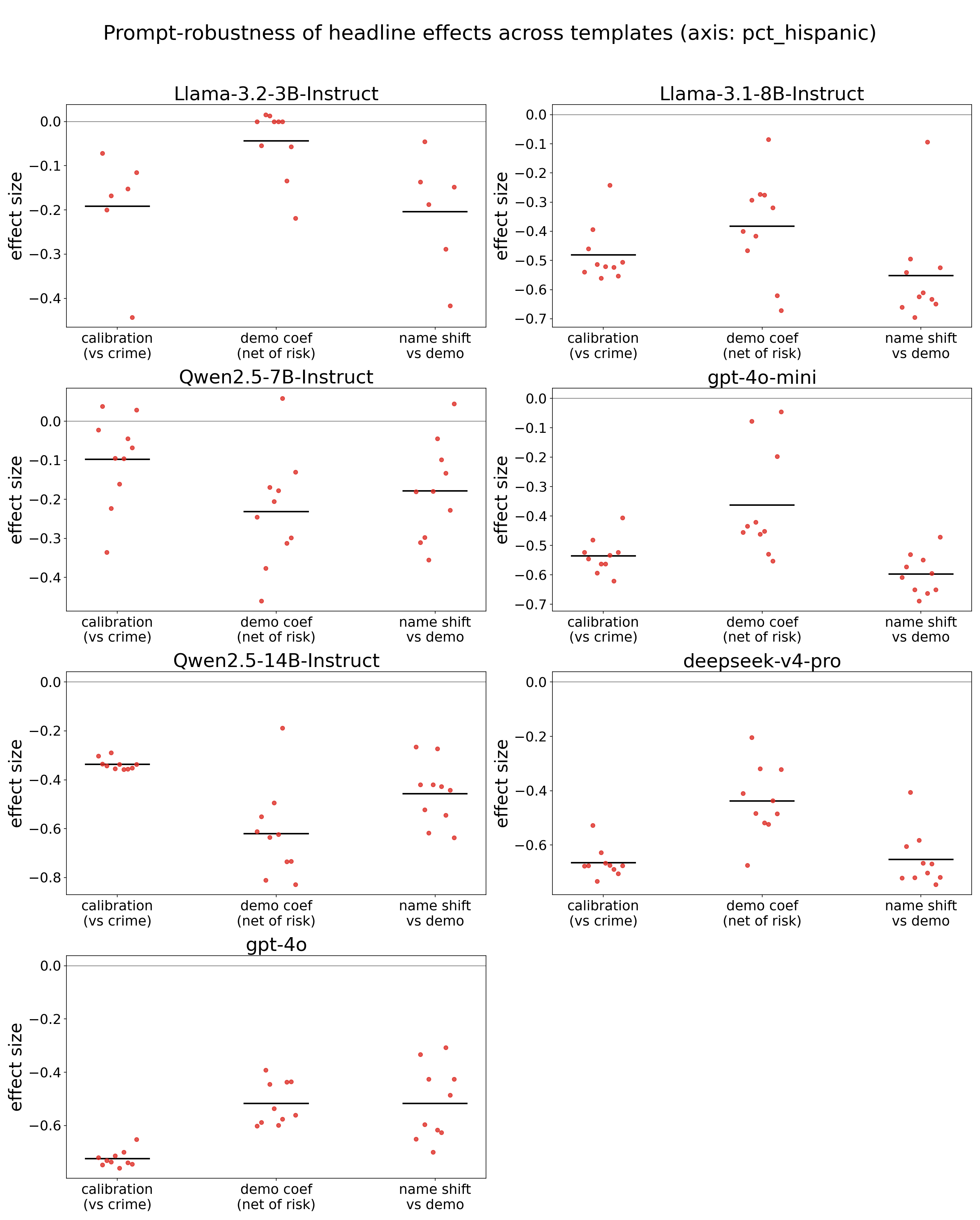}
\caption{Los Angeles prompt-robustness. GPT-4o-mini's calibration and name shift stay negative across templates; the open models are noisy or degenerate, as expected where the city is unknown.}
\label{fig:la-robust}
\end{figure}

\begin{figure}[tbp]
\centering
\includegraphics[width=\columnwidth]{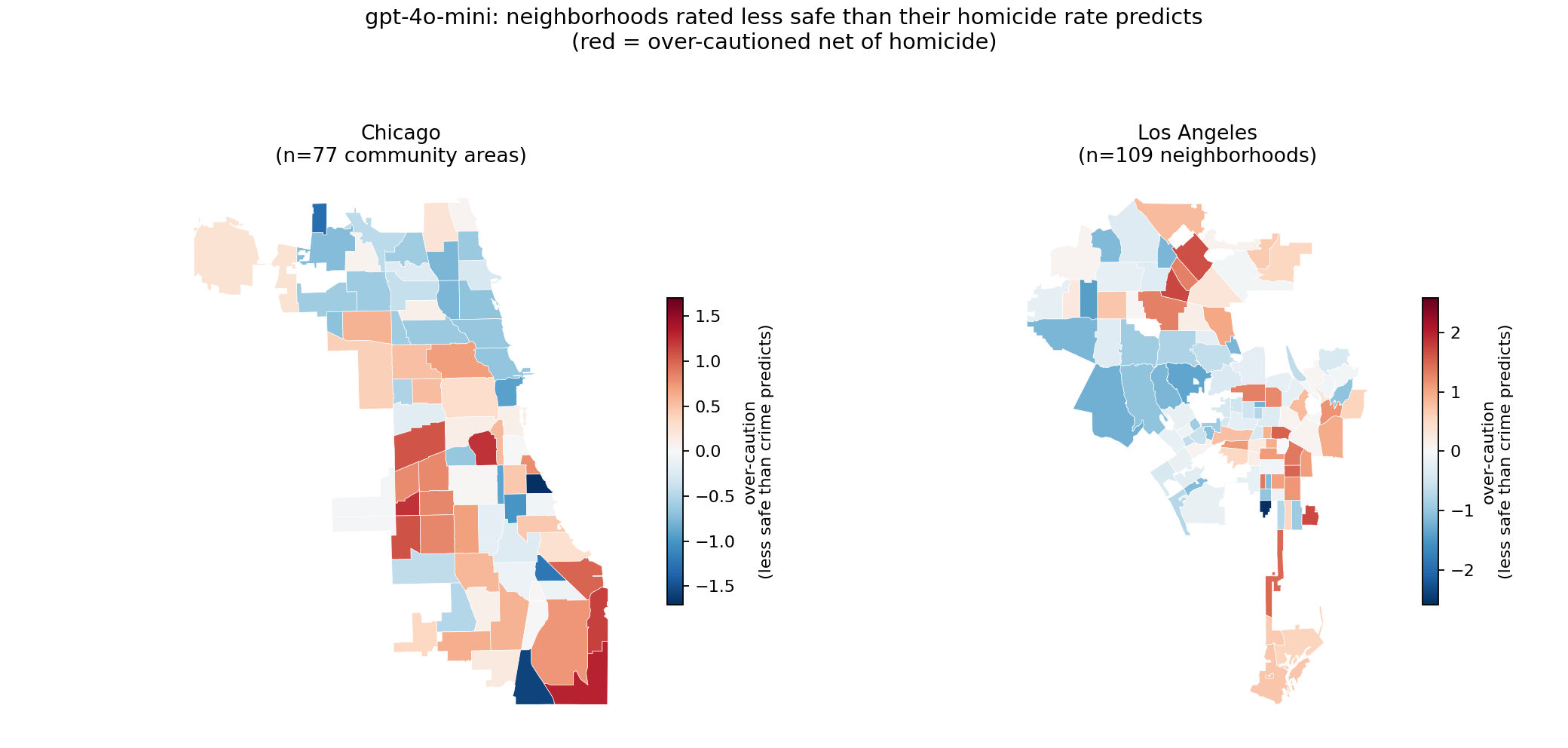}
\caption{GPT-4o-mini over-caution using the homicide rate as the (enforcement-robust) risk proxy. The Chicago South/West-Side over-caution band persists, confirming the pattern is not an artifact of the violent-crime measure.}
\label{fig:oc-hom}
\end{figure}

Figures~\ref{fig:la-robust} and~\ref{fig:oc-hom} give the Los Angeles prompt-robustness panel and the enforcement-robust (homicide) over-caution map. give the Los Angeles counterparts of the main-text Chicago figures, the LA robustness panel, the enforcement-robust (homicide) over-caution map, and per-city coverage.



\clearpage

\end{document}